\documentclass[letterpaper,twocolumn,fleqn]{article} 

\usepackage{ist}
\usepackage{multirow} 
\usepackage{comment}
\usepackage[hidelinks]{hyperref}
\usepackage{tabularx}
\usepackage{colortbl} 
\usepackage{xcolor}
\definecolor{lightgray}{rgb}{0.83, 0.83, 0.83}

\usepackage{caption}
\title{Revisiting Birds Eye View Perception Models with Frozen Foundation Models: DINOv2 and Metric3Dv2}

\author{Seamie Hayes$^{1,2,3}$, Ganesh Sistu$^{1,3,4}$, Ciaran Eising$^{1,2,3}$\\
$^1$Dept. of Electronic and Computer Engineering, University of Limerick, Castletroy, Co. Limerick  V94 T9PX, Ireland\\
$^2$SFI CRT Foundations in Data Science, University of Limerick, Castletroy, Co. Limerick  V94 T9PX, Ireland\\
$^3$Data Driven Computer Engineering (D²iCE) Research Centre, University of Limerick, Castletroy, Co. Limerick  V94 T9PX, Ireland\\
$^4$Valeo Vision Systems, Tuam, Galway DY1 22DJ, Ireland
\vspace{-1cm}}
\date{} 

\begin{document} 

\maketitle 

\thispagestyle{empty} 


\begin{abstract}
    Birds Eye View perception models require extensive data to perform and generalize effectively. While traditional datasets often provide abundant driving scenes from diverse locations, this is not always the case. It is crucial to maximize the utility of the available training data. With the advent of large foundation models such as DINOv2 and Metric3Dv2, a pertinent question arises: can these models be integrated into existing model architectures to not only reduce the required training data but surpass the performance of current models? We choose two model architectures in the vehicle segmentation domain to alter: Lift-Splat-Shoot, and Simple-BEV. For Lift-Splat-Shoot, we explore the implementation of frozen DINOv2 for feature extraction and Metric3Dv2 for depth estimation, where we greatly exceed the baseline results by 7.4 IoU while utilizing only half the training data and iterations. Furthermore, we introduce an innovative application of Metric3Dv2's depth information as a PseudoLiDAR point cloud incorporated into the Simple-BEV architecture, replacing traditional LiDAR. This integration results in a +3 IoU improvement compared to the Camera-only model.
\end{abstract}

\section{Introduction}
The recent revolution in machine learning has paved the way for the creation of numerous vehicle segmentation networks for use in autonomous vehicles, all competing to achieve optimal performance. The prompt yet accurate detection of vehicles in an image is no easy task, and the utilization of large convolutional networks, or transformer models has been extensively investigated. To train these networks effectively, they require dense information from which to learn, typically derived from three sensor modalities—camera, LiDAR, and radar—used individually or in combination. These modalities are complementary: cameras deliver rich 2D colour images that enhance vehicle segmentation and lane detection. LiDAR and radar generate point clouds, significantly improving vehicle segmentation and depth perception. When integrated with camera data, they outperform models that rely solely on camera input \cite{simplebev}.

Training autonomous perception models can be time-consuming due to the extensive data requirements and the size of datasets like nuScenes \cite{nuScenes}, which includes 700 driving scenes each with 40 annotated keyframes. The Lift-Splat-Shoot (LSS) model, previously considered state-of-the-art, is trained for approximately 40 epochs on this dataset and achieves a performance of approximately 33 IoU. Reducing training time while maintaining performance would be remarkable, and this would require a readily available model pre-trained on a massive dataset to replace a particular task in the architecture: a foundation model. A foundation model is trained on a large, diverse dataset, for the purpose of generalisation and then applicable to tasks such as image object segmentation, depth perception, text classification, etc. The advancement of foundation models cannot be downplayed. For instance, EfficientSAM \cite{efficientSAM} processes images significantly faster than the original SAM \cite{sam}, with minimal performance loss. Likewise, for depth perception, Depth Pro \cite{depthpro} outperforms Metric3Dv2 \cite{metric3dv2} in monocular depth estimation, producing a 2.25-megapixel depth map in just 0.3s.

We chose two autonomous bird's eye view vehicle segmentation models for modification: Lift-Splat-Shoot (LSS) and Simple-BEV. LSS, a well-established model, shows strong performance in vehicle segmentation. In contrast, Simple-BEV, which utilizes LiDAR and radar data from nuScenes, outperforms LSS. For LSS, we implemented foundational models: DINOv2 for feature extraction \cite{dinov2} and Metric3Dv2 for depth estimation, using them in their frozen state to avoid fine-tuning and preserve original weights. It's crucial to note that DINOv2 is designed to capture all image features, not just vehicles. For Simple-BEV, we experimented with Metric3Dv2 depth images as a PseudoLiDAR point cloud, aiming to replace traditional LiDAR or radar data. Results demonstrate that these foundational models significantly enhance initial performance due to their versatility. By using only camera-based solutions, which are significantly cheaper than LiDAR, there's potential for substantial cost savings in equipment manufacturing.

The main contributions of this paper are as follows:
\begin{enumerate}
  \item The implementation of foundation models in Lift-Splat-Shoot significantly enhances results, even with only half the training data and fewer iterations. Additionally, an ablation study examines how the size of foundation models affects performance.
  \item The unique implementation of metric depth images to simulate LiDAR in Simple-BEV which yields better results than the Camera-only model.
\end{enumerate}

\section{Background}
\subsection{Lift-Splat-Shoot}

\begin{figure}[!ht] 
  \begin{center}
    \includegraphics[width=\columnwidth]{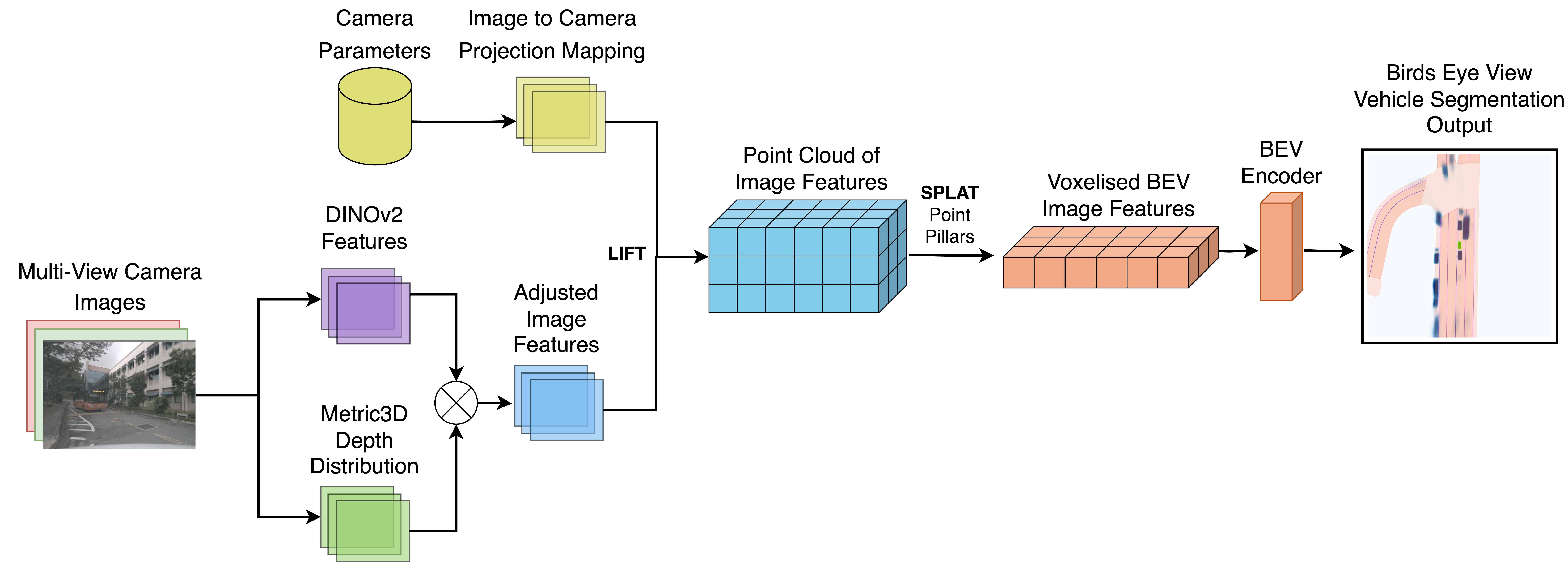}
  \end{center}
  \caption{\textbf{Architecture Flow Diagram: }Modified Lift-Splat architecture with implementation of DINOv2 and Metric3Dv2 for image feature extraction and depth estimation}
  \label{arch_new_lss}
\end{figure}

The LSS model consists of three primary components. The first, lifting, transforms each image into a frustum of features using a geometry transformation and an EfficientNet backbone for acquiring image features and depth estimation \cite{liftsplat}. Next, the splat step encodes these frustums into a BEV voxel grid using PointPillars, which is then processed by a BEV encoder to produce the BEV vehicle segmentation output. The final component, shoot, involves trajectory planning for the ego vehicle, selecting the path with the lowest cost determined by a network. This paper focuses primarily on the Lift component, which pertains to the acquisition of camera features and depth estimation. Here, replace the output of the EfficientNet encoder with DINOv2 image features and Metric3Dv2 depth to facilitate increased model performance, which will be described later in Section \ref{sec:alt_lss}. This modified model architecture is illustrated in Figure \ref{arch_new_lss}

\subsection{Simple BEV}
\begin{figure}[!ht] 
  \begin{center}
    \includegraphics[width=\columnwidth]{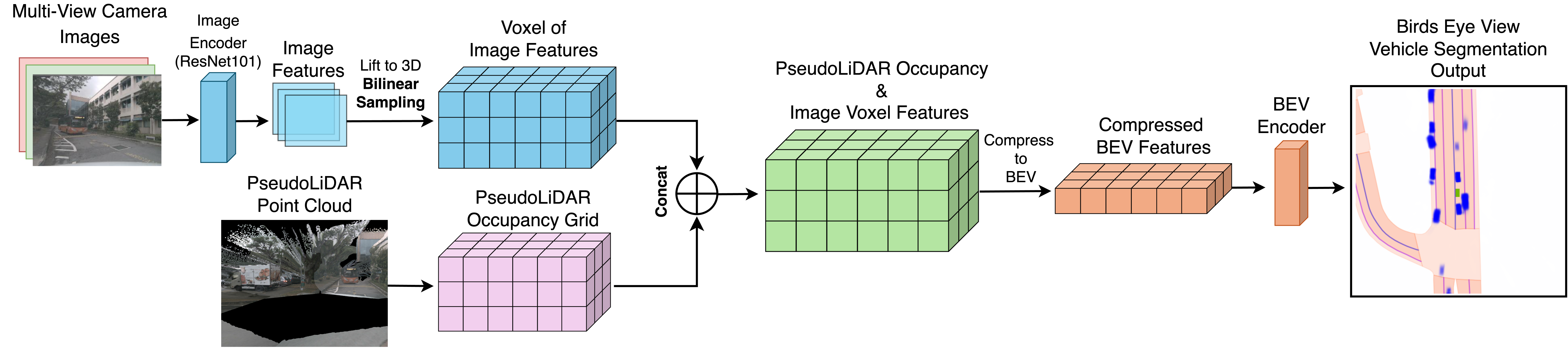}
  \end{center}
  \caption{\textbf{Architecture Flow Diagram: }Simple BEV  model for Camera+PseudoLiDAR. The Camera-only model excludes any point clouds, while Radar and LiDAR would substitute for Pseudo LiDAR}
  \label{arch_bev}
\end{figure}

The Simple BEV architecture allows us to implement radar or LiDAR data along with camera information which makes it a very versatile model. We extract features from each image and then perform bilinear sampling to lift these features into a 3D voxel. For LiDAR and radar data, we convert the point cloud information into an occupancy grid representation. We then concatenate the two voxels together to yield a unified voxel, compress it to the BEV space, and input it to the BEV encoder, producing our BEV vehicle segmentation output. A comprehensive flow diagram is provided in Figure \ref{arch_bev}. 

Additionally, we have the option to train three models: Camera-only, Camera+Radar, or Camera+LiDAR. LiDAR and radar provide the model with important depth information and also permit the model to segment vehicles not seen by the camera images due to the range of these sensors. Both modalities provide notable performance increases over the single-modal Camera-only model.

\subsection{DINOv2}
The significance of foundation models in computer vision cannot be overstated, especially with the advent of DINOv2 \cite{dinov2}. Trained on over 140 million images from diverse datasets, this self-supervised model leverages a Vision Transformer-based backbone to excel in zero-shot inference. By dividing images into $14\times14$ patches, DINOv2 effectively captures long-range dependencies, allowing even its frozen version to outperform fine-tuned conventional models, as detailed in Section \ref{sec:lss_comp}. The model scales feature vector length from 384 in smaller versions to larger dimensions in bigger variants. While fine-tuning DINOv2 for specific tasks is feasible, it requires significant computational resources. In our study, we employ the frozen model to sidestep these complexities, incorporating a single convolutional layer to downsample DINOv2 feature vectors to 64, thus ensuring compatibility with the EfficientNet BEV encoder. As shown in Table \ref{tab:model_param}, DINOv2's extensive parameter set and robust pre-training regimen set it apart from standard CNN architectures.

\begin{table}[!ht]
\centering
\begin{tabular}{ccc}
    \hline
    \textbf{\textit{Model}} & \textbf{\textit{Parameters}} & \textbf{\textit{Dataset Size}}\\
    \hline
    DINOv2 Small & 21M & 142M\\
    DINOv2 Base & 86M & 142M\\
    DINOv2 Large & 300M & 142M\\
    DINOv2 Giant & 1,100M & 142M\\
    \hline \hline
    EfficientNet-b0 & 5.3M & 1.2M\\
    ResNet-101 & 44.5M & 1.2M\\
    \hline
\end{tabular}
\vspace{1mm}
\caption{\textbf{Model Comparison}: Various DINOv2 variants compared to EfficientNet-b0 and ResNet-101 used in Lift-Splat-Shoot and Simple-BEV respectively. Dataset Size refers to the number of images the model is pre-trained on.}
\label{tab:model_param}
\end{table}

\subsection{Metric3Dv2}
Metric3Dv2 is a zero-shot single-view metric depth model that utilizes a canonical camera space to overcome the metric ambiguity typically associated with standard monocular cameras \cite{metric3dv2}. By transforming monocular images to canonical camera space and being trained on over 8 million images from thousands of different cameras, Metric3Dv2 generalizes effectively to provide quality pixel-level meter depth. In our experiments, we will utilise this depth information to aid the projection of image features into 3D and, in addition to this, create a 3D point cloud for emulating LiDAR.
 
\section{Implementation of Foundation Models in Lift-Splat-Shoot} \label{found_in_lss}
\subsection{Alteration to the Model Architecture}
\label{sec:alt_lss}
One of the main contributions of this paper is the implementation of DINOv2 features and Metric3Dv2 metric depth maps to the LSS architecture. As explained previously, LSS uses an EfficientNet encoder, which produces feature maps and depth distributions of images in an effort to efficiently predict the precise location of vehicles in 3D space. We will replace the EfficientNet backbone with our two foundation models: DINOv2 and Metric3Dv2. One important note is that EfficientNet downsamples the image by a factor of 16, and DINOv2 downsamples by a factor of 14. Therefore, we now have more information from DINOv2 due to the smaller patches, which allows for finer granularity in feature extraction compared to EfficientNet. The changed architecture flow is illustrated in Figure \ref{arch_new_lss}

\subsection{Metric Depth Distribution}
\begin{figure}[!ht] 
  \vspace{-0.4cm}
  \begin{center}
    \includegraphics[width=\columnwidth]{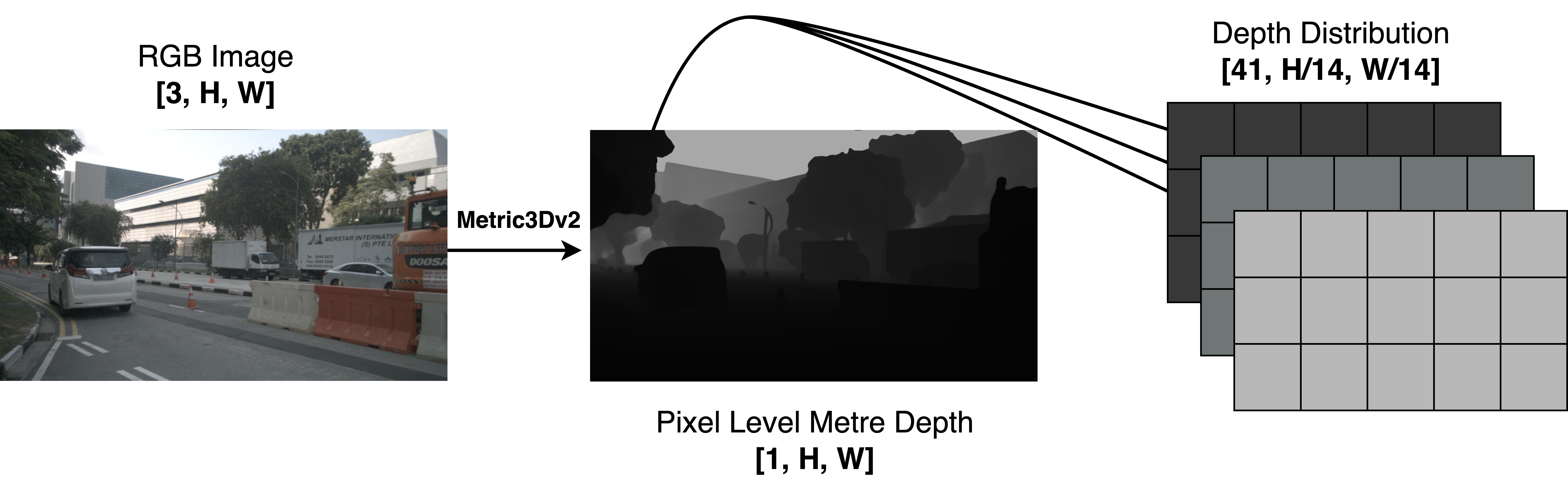}
  \end{center}
  \vspace{-0.25cm}
  \caption{\textbf{Metric3Dv2: }Conversion from RGB image to depth image to depth distribution for compatibility with DINOv2 patch embeddings}
  \label{metric}
\end{figure}

Metric3Dv2 provides pixel-level depth in meters. However, in the LSS architecture, the depth distribution is divided into 41 bins, representing uniform depths ranging from 4 to 45 meters. In addition to this, the depth distribution is not represented on a pixel-by-pixel basis but instead in a 16$\times$16 patch manner. Hence, our Metric3Dv2 depth distribution must match the format used in the model. Here, we present a method of converting the continuous depth image into a discrete depth distribution. To transform our depth image accordingly, we pool our depths using a non-overlapping 16$\times$16 square across the depth map. For each pixel contained within this patch, we pool its value into the corresponding meter bin of our depth distributions. This process converts a depth image of size $[1, H, W]$ and transforms it to size $[41, H/16, W/16]$ as required, illustrated in Figure \ref{metric}. For use with DINOv2, we downsample by 14 to match its downsampling factor. For training stability and increased model performance, we apply one same-convolutional layer with batch normalization and ReLU. The final result is a tensor representing the probability of a DINOv2/EfficientNet patch being present at each meter bin depth, exactly as the original model intended.

For the model architecture described in Figure \ref{arch_new_lss}, we will apply it to the full nuScenes training dataset (700 scenes) and half of the dataset (350 scenes). For inference, we retain the 150 validation scenes. We will compare our modified model to the original model performance on these datasets while also performing an ablation utilizing varying DINOv2 and Metric3Dv2 model sizes.

\subsection{Results: Largeness of Foundation Models}
In this section, we implement DINOv2 and Metric3Dv2 in LSS and analyze the effect the size of these models has on performance.

\begin{table}[!ht]
\centering
\begin{tabular}{cccc}
    \hline
    \multicolumn{2}{c}{\textbf{Configuration}} & \multicolumn{2}{c}{\textbf{Peak}} \\
    \hline
    \textbf{\textit{DINOv2 Size}} & \textbf{\textit{Metric3Dv2 Size}} & \textbf{\textit{IoU}} & \textbf{\textit{Iterations}} \\
    \hline
    Giant & Small & 36.1 & 50k \\
    Giant & Large & 39.6 & 80k \\
    Giant & Giant & \textbf{41.9} & 55k \\
    \hline \hline
    Small & Giant & 40.1 & 120k \\
    Base & Giant & 41.0 & 145k \\
    Large & Giant & 41.1 & 125k \\
    Giant & Giant & \textbf{41.9} & 55k \\
    \hline
\end{tabular}
\vspace{1mm}
\caption{\textbf{Comparing performance versus size of foundation models for LSS}: Largeness of models affects performance. All models are trained on the full dataset for 250,000 iterations}
\label{tab:iou_large_dino_metric}
\end{table}

First, we replace the feature extractor with DINOv2 and the depth estimator with Metric3Dv2.  When analyzing one foundation model, we will keep the other model fixed at the Giant state. By examining Table \ref{tab:iou_large_dino_metric}, we observe that the size of the DINOv2 and Metric3Dv2 models significantly impacts performance. This is especially evident with Metric3Dv2, where implementing the Giant model results in an IoU improvement of approximately 5.8 compared to the small model. In contrast, DINOv2 shows a smaller improvement, with only a 1.8 IoU increase when using the Giant model over the small model. Additionally, these models reach peak performance relatively quickly, all in under 150k iterations, which is considerably less than the LSS model, which requires 300k+ iterations to train fully. We place minimal emphasis on iterations when comparing these models, primarily because of their rapid convergence; typically, these models achieve performance within ±0.5 IoU of the peak value within approximately 50,000 iterations.

\begin{table}[!ht]
\centering
\begin{tabular}{cccc}
    \hline
    \multicolumn{2}{c}{\textbf{Configuration}} & \multicolumn{2}{c}{\textbf{Peak}} \\
    \hline
    \textbf{\textit{Feature Extractor}} & \textbf{\textit{Metric3Dv2 Size}} & \textbf{\textit{IoU}} & \textbf{\textit{Iterations}} \\
    \hline
    EffNet & Small & 34.3 & 315k \\
    EffNet & Large & 38.6 & 150k \\
    EffNet & Giant & \textbf{40.5} & 110k \\
    \hline
\end{tabular}
\vspace{1mm}
\caption{\textbf{Comparing performance versus size of Metric3Dv2 model for depth estimation}: Largeness affects performance}
\label{tab:iou_large_metric3d}
\end{table}

Secondly, we will only replace the depth estimator with Metric3Dv2, leaving feature extraction to EfficientNet. We observe similar results, wherein the Giant model significantly outperforms the Small model, exhibiting an increase of 6.2 IoU seen in Table \ref{tab:iou_large_metric3d}. An interesting result is seen when comparing the DINOv2+Metric3Dv2 (41.9 IoU) versus EffNet+Metric3Dv2 (40.5 IoU). From this, we infer that Metric3Dv2 contributes more to performance improvement than DINOv2 does. DINOv2 provides an initial boost in performance. However, EfficientNet nearly approaches this performance given sufficient training time.

Utilizing DINOv2 for feature extraction and EfficientNet for depth estimation is possible but omitted from this paper due to the discrepancy in downsampling for both models. This would require different-sized images for both models to achieve the same number of image patches, leading to an unfair experiment.

\subsection{Results: Comparison with Original Model}
\label{sec:lss_comp}

In this section, we will compare the performance of our altered LLS architecture directly with the original LSS model.

\begin{table}[!ht]
\centerline{%
\renewcommand{\arraystretch}{1.2} 
\begin{tabular}{cccccc}
    \hline
    \multicolumn{4}{c}{\textbf{Configuration}} & \multicolumn{2}{c}{\textbf{Peak}} \\
    \hline
    \parbox[c][1.3cm][c]{1.35cm}{\centering \vspace{0.5em} \textbf{\textit{Feature} \\ \textit{Extractor}} \vspace{0.5em}} & 
    \parbox[c][1.3cm][c]{1.35cm}{\centering \vspace{0.5em} \textbf{\textit{Depth} \\ \textit{Extractor}} \vspace{0.5em}} & 
    \parbox[c][1.3cm][c]{0.9cm}{\centering \vspace{0.5em} \textbf{\textit{Model} \\ \textit{Size}} \vspace{0.5em}} & 
    \parbox[c][1.3cm][c]{0.9cm}{\centering \vspace{0.5em} \textbf{\textit{Training} \\ \textit{Data}} \vspace{0.5em}} & 
    \parbox[c][1cm][c]{1cm}{\centering \vspace{0.5em} \textbf{\textit{IoU $\uparrow$}} \vspace{0.5em}} & 
    \parbox[c][1.1cm][c]{0.7cm}{\centering \vspace{0.5em} \textbf{\textit{Iters}} \vspace{0.5em}} \\
    \hline
    \rowcolor{lightgray} 
    EffNet & EffNet & N/A & Full & 33.0 & 300k \\

    \hline
    DINOv2 & Metric3Dv2 & Giant & Full & \textbf{41.9} & 55k \\
    DINOv2 & Metric3Dv2 & Small & Full & 34.0 & 115k \\
    \hline
    EffNet & Metric3Dv2 & Giant & Full & 40.5 & 110k \\
    EffNet & Metric3Dv2 & Small & Full & 34.1 & 115k \\
    \hline \hline 
    \rowcolor{lightgray} 
    EffNet & EffNet & N/A & Half & 29.1 & 150k \\
    \hline
    DINOv2 & Metric3Dv2 & Giant & Half & \textbf{40.4} & 35k \\
    DINOv2 & Metric3Dv2 & Small & Half & 32.1 & 30k \\
    \hline
    EffNet & Metric3Dv2 & Giant & Half & 37.4 & 55k \\
    EffNet & Metric3Dv2 & Small & Half & 31.4 & 80k \\
    \hline
\end{tabular}
}
\vspace{1mm}
\caption{\textbf{IoU Comparison for Various LSS Architectures}: Baseline model results are highlighted in grey. DINOv2 and Metric3Dv2 provide increased model performance with reduced training data}
\label{tab:iou_comp}
\end{table}

First we train on the full dataset, and as shown in Table \ref{tab:iou_comp}, foundation models demonstrate significant improvements. The Giant foundation models greatly outperform the original model by 8.9 IoU. Moreover, these Giant models surpass the original model by 4.8 IoU after 5k training iterations, which is considerably small compared to the 150k iterations to train the original model. Similar results are observed with the Small foundation models, which surpass the original model after 30k iterations. Despite decreased performance compared to their Giant counterparts, the Small foundation models may be preferable in certain scenarios due to decreased training time. For Metric3Dv2+EffNet, we find that higher resolution depth information provides the most significant enhancement, yielding a 7.5 IoU increase compared to the original model. Additionally, it surpasses the performance of the original model after 5k iterations.

Finally, we re-train all models using only half of the training data. The original LSS model's performance dropped to 29.1 IoU, a decrease of 4.1 IoU. Results are observed similarly to the full dataset training, where for the Giant variation, we see great improvement, even surpassing the original LSS model trained on the full dataset. Our DINOv2+Metric3Dv2 Giant models outperform the original full dataset model by 7.4 IoU. However, we observe that the small variations of the models fail to outperform the original LSS model trained on the full dataset. We conclude that foundation models bring notable gains in performance when utilized on old model architectures, even when utilizing just half the training data.

A significant limitation is the runtime: during training the Giant model processes 1,000 iterations in 900 seconds, whereas the original model only takes 120 seconds, likely due to the increased size of the Giant model. Additionally, our Metric3Dv2 depth images are pre-calculated, requiring about 1 second per image. Without pre-calculation, the runtime would increase substantially. All models are trained on one NVIDIA A100-SXM4-40GB.

\section{Implementation of Foundation Models in Simple-BEV}
Following the previous experiments, we will attempt to implement these models on a newer architecture. We note the implementation of DINOv2 in Simple-BEV has been explored and aids in achieving similar results to the original model while taking two-thirds fewer iterations and significantly reduced model parameters due to the use of a low-rank approximation of the DINOv2 embeddings \cite{lora, dinov2_sbev}. One important note is that utilizing the frozen DINOv2 model yields poor results compared to the original Simple-BEV model. 

\subsection{Construction of a PseudoLiDAR Point Cloud with Metric3Dv2}
One method to implement Metric3Dv2 depth in Simple-BEV is by replacing the bilinear sampling voxelization method with depth-based splatting. However, referring to the Simple-BEV paper, depth-based splatting yields an IoU loss of -3 IoU \cite{simplebev}. We propose a method that preserves the bilinear sampling splatting technique while effectively leveraging the informative depth images produced by Metric3Dv2. Depth information such as LiDAR or radar provides a considerable boost in performance over Camera-only models, +13.4 for LiDAR and +8.3 for radar. Taking inspiration from this, we convert our depth images into a unified point cloud by projecting them into 3D space.

\begin{figure}[!ht] 
  \begin{center}
    \includegraphics[width=\columnwidth]{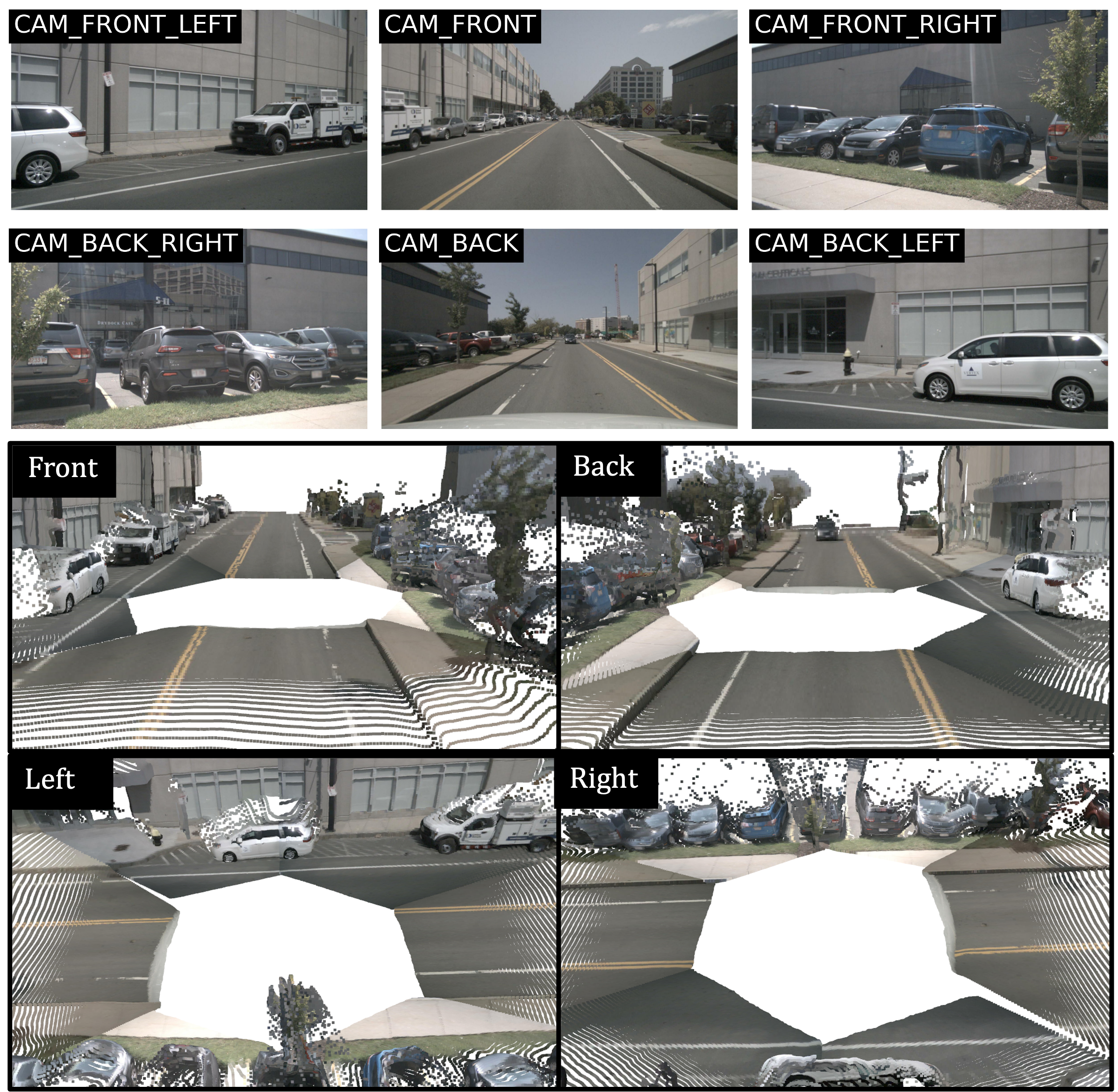}
  \end{center}
  \vspace{-0.25cm}
  \caption{\textbf{Metric3Dv2 Point Cloud Sample: }PseudoLiDAR point cloud extracted from a nuScenes sample at a depth image size of $112\times200$). RGB is coupled for visualisation purposes}
  \label{metric_sbev_sample}
\end{figure}

We first obtain a depth image from our original image from Metric3Dv2; then, we generate a 3D point cloud from this 2D depth image using the intrinsic and extrinsic camera information. The density of our point cloud inherently depends on the resolution of our depth image. For our experiments, we limit our depth image resolution to $112\times224$ and $224\times400$ to save on computation costs, although we use the Giant Metric3Dv2 for maximal quality of the depth maps. A visualisation of this point cloud is provided in Figure \ref{metric_sbev_sample}.

Some benefits of our Pseudo LiDAR point cloud can be observed over LiDAR. Our PseudoLiDAR point cloud requires only one sweep for effective performance, reducing computational and data loading demands compared to the five sweeps needed for optimal radar and LiDAR results. It also syncs directly with camera images, unlike the nuScenes sensors. Furthermore, even with low-resolution images, our approach generates a dense point cloud comparable to about three LiDAR sweeps.

\subsection{Results: Comparison with Original Model}

\begin{table}[!ht]
\centering
\renewcommand{\arraystretch}{1.1} 
\begin{tabular}{ccc}
    \hline
    \multicolumn{2}{c}{\textbf{Configuration}} & \textbf{Peak} \\
    \hline
    \textbf{\textit{Modality}} & \textbf{\textit{Depth Image Size}} & \textbf{\textit{IoU}} \\
    \hline
    Camera & N/A & 47.4 \\
    Camera+Radar & N/A & 55.7 \\
    Camera+LiDAR & N/A & 60.8 \\
    \hline \hline
    Camera+PseudoLiDAR (ours) & (112,200) & \textbf{50.3} \\
    Camera+PseudoLiDAR (ours) & (224,400) & \textbf{50.7} \\
    \hline
\end{tabular}
\vspace{1mm}
\caption{\textbf{Performance Comparison of baseline Simple-BEV models versus our Camera+PseudoLiDAR model:} Trained for 25k iterations and input image size of $448\times800$}
\label{tab:sbev_metric}
\end{table}

From Table \ref{tab:sbev_metric}, we examine the gains this metric depth information provides over the Camera-only model, increasing by approximately +3 IoU. We also see utilizing larger-depth image resolutions yields an insignificant increase in IoU, showing that low-depth image quality is sufficient. This demonstrates that Metric3Dv2 supplies valuable depth information, which enhances the model's ability to segment vehicles more effectively.

Despite the higher density of our PseudoLiDAR point cloud—threefold that of LiDAR—it does not outperform the latter. This limitation arises because the point cloud replicates the camera image data in a depth format. Given enough training, the base Camera-only model capably infers this depth through bilinear sampling. This becomes apparent when training at a resolution of $224\times400$, which is half that of the original image as evident in Table \ref{tab:sbev_metric_res1}, where we see an IoU improvement of +6.4, much greater than the improvement of +3.0 of the full image resolution. Here, our point cloud provides useful information that the model failed to learn from the image features alone.

\begin{table}[!ht]
\centering
\renewcommand{\arraystretch}{1.1} 
\begin{tabular}{ccc}
    \hline
    \multicolumn{2}{c}{\textbf{Configuration}} & \textbf{Peak} \\
    \hline
    \textbf{\textit{Modality}} & \textbf{\textit{Depth Image Size}} & \textbf{\textit{IoU}} \\
    \hline
    Camera & N/A & 42.3 \\
    Camera+PseudoLiDAR (ours) & (112,200) & \textbf{48.6} \\
    \hline
\end{tabular}
\vspace{1mm}
\caption{\textbf{Performance comparison of Simple-BEV Camera-only model vs. Camera+PseudoLiDAR model:} Trained for 25k iterations and input image size of $224\times400$}
\label{tab:sbev_metric_res1}
\end{table}

Additionally, the PseudoLiDAR model provides an initial boost in performance. For the Camera-only model at an image resolution of $224\times400$, we see a 28.8\% improvement in IoU from iteration 2,500 to 25,000. However, for Camera+PseudoLiDAR, we see an improvement of just 20.3\%, despite a higher final IoU over camera-only. Additionally, we attempted to decorate our point cloud with RGB metadata, but this worsened its performance.

\section{Qualitative Analysis: BEV Vehicle Segmentation}

Analysing IoU provides a near-comprehensive outlook on the model's performance across the entire validation dataset. However, it is still necessary to examine the model's visual output to seek anomalies in its detections. It is important to minimize Type II errors when deploying these models in autonomous vehicles.

\subsection{Lift-Splat-Shoot}

\begin{figure}[htbp] 
  \begin{center}
    \includegraphics[width=\columnwidth]{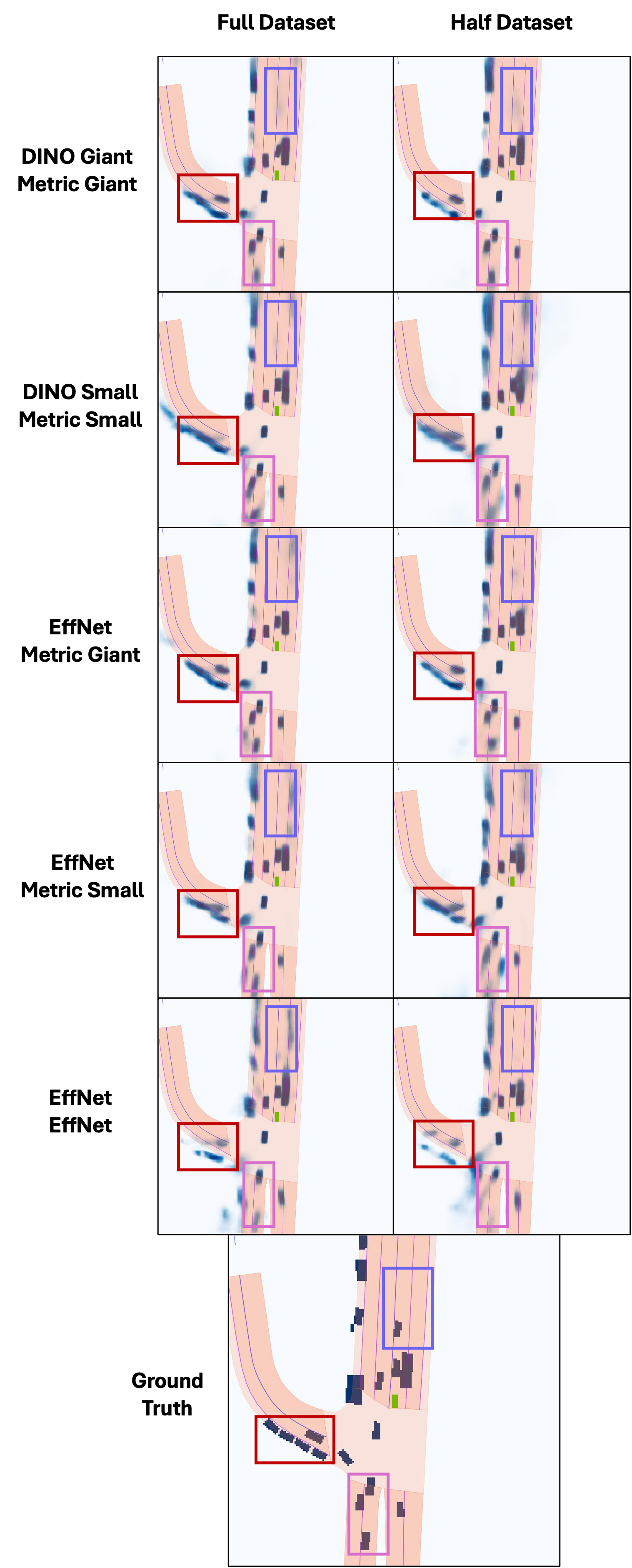}
  \end{center}
  \caption{\textbf{Lift-Splat-Shoot: }Comparison for full dataset and half dataset models. Vehicles are highlighted by coloured boxes for the purpose of analysis}
  \label{lss_vis}
\end{figure}

Upon examining Figure \ref{lss_vis}, a notable distinction is seen between the outputs of the full and half dataset models. The half-dataset model exhibits a 'ghosting' effect around vehicles due to model uncertainty in predicting exact position. As we progress upward in the figure, there is a noticeable improvement in vehicle segmentation quality resulting from the integration of foundation models into the architecture.

The pink-highlighted vehicles are poorly segmented by the original model, which also inaccurately predicts vehicles outside the drivable area. Implementing DINOv2 and Metric3Dv2 increases certainty, particularly for the Giant variations. The red-highlighted vehicles are challenging for all models, but DINOv2 and Metric3Dv2 enhance their segmentation, likely aided by superior depth information. Conversely, in the blue-highlighted area, the 'ghosting' issue persists with the original model predicting non-existent vehicles. All models face difficulties accurately segmenting this vehicle, primarily due to occlusion by other vehicles.

\subsection{Simple-BEV}
\begin{figure}[!ht]
  \vspace{-3mm}
  \begin{center}
    \includegraphics[width=\columnwidth]{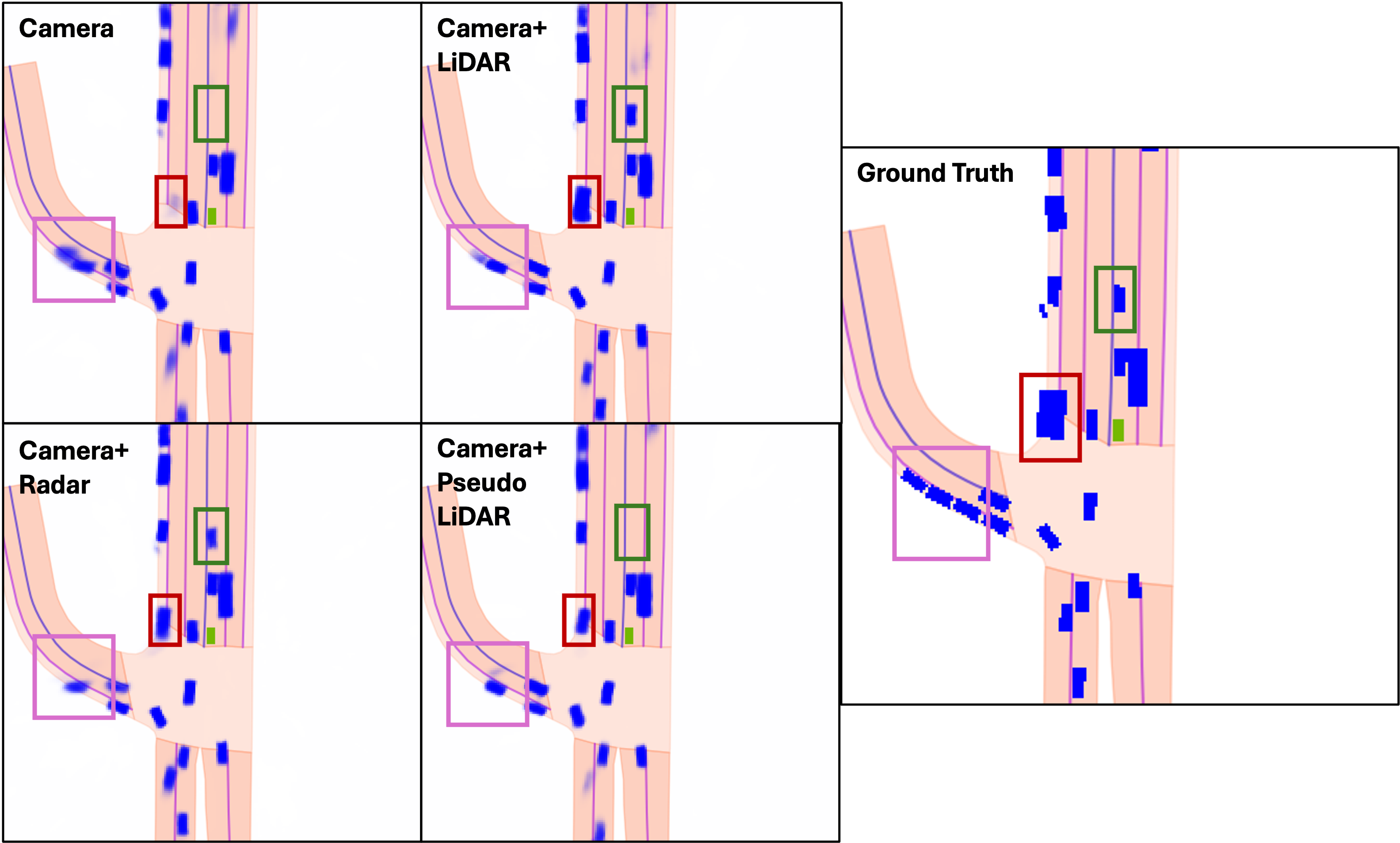}
  \end{center}
  \caption{\textbf{Simple-BEV Comparison: }Camera only vs Camera+Radar vs Camera+LiDAR vs Camera+Pseudo LiDAR. Vehicles are highlighted by coloured boxes for the purpose of analysis}
  \label{sbev_vis}
\end{figure}

Examining Figure \ref{sbev_vis} we examine the improvement depth information provides over Camera-only. The red-highlighted vehicle is segmented only when depth is implemented, as the Camera-only model fails here. Similarly, the green-highlighted vehicle is segmented by the Camera+radar and Camera+LiDAR models, showing that multiple sensors enhance vehicle detection by providing diverse, complementary data.

The pink-highlighted vehicles are most accurately segmented by the Camera-only model, whereas they are not identified by the other models that also incorporate camera input. This discrepancy may arise because radar or LiDAR data did not detect these vehicles, potentially causing them to be misclassified as false positives. Although the PseudoLiDAR point cloud substantially enhances segmentation compared to the Camera-only model, its performance still falls short when compared to multi-modal models.

\section{Conclusion}
These experiments show that foundation models like DINOv2 and Metric3Dv2 significantly boost performance when integrated into Lift-Splat-Shoot, and reduce training data/iterations required. Using Metric3Dv2 for our PseudoLiDAR point cloud yields notable improvements over Simple-BEV's Camera-only model. The primary performance bottlenecks are the quality of the feature extractor and the voxelization process, highlighting the need for focused improvements in these areas. We hope this study, along with others on the role of foundation models in BEV models, inspires further research in this field. 

\section*{Acknowledgements}
This publication has emanated from research conducted with the financial support of Science Foundation Ireland under Grant number 18/CRT/6049. For the purpose of Open Access, the author has applied a CC BY public copyright license to any Author Accepted Manuscript version arising from this submission.

\begin{thebibliography}{99}
    \bibitem{simplebev} Harley, A. W., Fang, Z., Li, J., Ambrus, R., and Fragkiadaki, K. (2023). Simple-BEV: What Really Matters for Multi-Sensor BEV Perception? In 2023 IEEE International Conference on Robotics and Automation (ICRA), pages 2759–2765.
    \bibitem{nuScenes} Caesar, H., Bankiti, V., Lang, A. H., Vora, S., Liong, V. E., Xu, Q., Krishnan, A., Pan, Y., Baldan, G., and Beijbom, O. (2020). nuScenes: A Multimodal Dataset for Autonomous Driving. In 2020 IEEE/CVF Conference on Computer Vision and Pattern Recognition (CVPR), pages 11618–11628.
    \bibitem{efficientSAM} Xiong, Y., Varadarajan, B., Wu, L., Xiang, X., Xiao, F., Zhu, C., Dai, X., Wang, D., Sun, F., Iandola, F., Krishnamoorthi, R., and Chandra, V. (2024). EfficientSAM: Leveraged Masked Image Pretraining for Efficient Segment Anything. In Proceedings of the IEEE/CVF Conference on Computer Vision and Pattern Recognition (CVPR), pages 16111–16121.
    \bibitem{sam} Kirillov, A., Mintun, E., Ravi, N., Mao, H., Rolland, C., Gustafson, L., Xiao, T., Whitehead, S., Berg, A. C., Lo, W.-Y., Dollar, P., and Girshick, R. (2023). Segment Anything. In Proceedings of the IEEE/CVF International Conference on Computer Vision (ICCV), pages 4015–4026.
    \bibitem{depthpro} Bochkovskii, A., Delaunoy, A., Germain, H., Santos, M., Zhou, Y., Richter, S. R., and Koltun, V. (2024). Depth Pro: Sharp Monocular Metric Depth in Less Than a Second. arXiv preprint arXiv:2410.02073.
    \bibitem{metric3dv2} Hu, M., Yin, W., Zhang, C., Cai, Z., Long, X., Chen, H., Wang, K., Yu, G., Shen, C., and Shen, S. (2024). Metric3Dv2 v2: A Versatile Monocular Geometric Foundation Model for Zero-shot Metric Depth and Surface Normal Estimation. IEEE Transactions on Pattern Analysis and Machine Intelligence, pages 1–18.
    \bibitem{dinov2} Pinetsuksai, N., Kittichai, V., Jomtarak, R., Jaksukam, K., Tongloy, T., Boonsang, S., and Chuwongin, S. (2023). Development of Self-Supervised Learning with Dinov2-Distilled Models for Parasite Classification in Screening. In 2023 15th International Conference on Information Technology and Electrical Engineering (ICITEE), pages 323–328
    \bibitem{liftsplat} Philion, J. and Fidler, S. (2020). Lift-Splat-Shoot: Encoding Images from Arbitrary Camera Rigs by Implicitly Unprojecting to 3D. In Computer Vision – ECCV 2020: 16th European Conference, Glasgow, UK, August 23–28, 2020, Proceedings, Part XIV, pages 194–210, Berlin, Heidelberg. Springer-Verlag.
    \bibitem{lora} Hu, E. J., Shen, Y., Wallis, P., Allen-Zhu, Z., Li, Y., Wang, S., Wang, L., and Chen, W. (2022). LoRA: Low-Rank Adaptation of Large Language Models. In International Conference on Learning Representations.
    \bibitem{dinov2_sbev} Barın, M. R., Aydemir, G., and Güney, F. (2024). Robust Bird's Eye View Segmentation by Adapting DINOv2. arXiv preprint arXiv:2409.10228.
    
\end{thebibliography}
\end{document}